\documentclass{article}

\usepackage{arxiv}

\usepackage[utf8]{inputenc} 
\usepackage[T1]{fontenc}    
\usepackage{hyperref}       
\usepackage{url}            
\usepackage{booktabs}       
\usepackage{amsfonts}       
\usepackage{nicefrac}       
\usepackage{microtype}      
\usepackage{cleveref}       
\usepackage{lipsum}         
\usepackage{graphicx}
\usepackage{natbib}
\usepackage{doi}
\usepackage{xcolor}
\usepackage{url}

\title{Computational Law: Datasets, Benchmarks, and Ontologies}

\newif\ifuniqueAffiliation
\uniqueAffiliationtrue

\author{Dilek K\"u\c{c}\"uk\thanks{Former legal intern at Ankara Bar Association.}\\
	T\"UB\.ITAK Marmara Research Center\\
	Ankara, Turkey\\
	\texttt{dilek.kucuk@tubitak.gov.tr} \\
	\And
	Fazli Can \\
	Bilkent University\\
	Ankara, Turkey\\
	\texttt{canf@cs.bilkent.edu.tr} \\
}

\renewcommand{\headeright}{}
\renewcommand{\undertitle}{}
\renewcommand{\shorttitle}{Computational Law: Datasets, Benchmarks, and Ontologies}

\date{}

\begin{document}
\maketitle

\begin{abstract}

Recent developments in computer science and artificial intelligence have also contributed to the legal domain, as revealed by the number and range of related publications and applications. Machine and deep learning models require considerable amount of domain-specific data for training and comparison purposes, in order to attain high-performance in the legal domain. Additionally, semantic resources such as ontologies are valuable for building large-scale computational legal systems, in addition to ensuring interoperability of such systems. Considering these aspects, we present an up-to-date review of the literature on datasets, benchmarks, and ontologies proposed for computational law. We believe that this comprehensive and recent review will help researchers and practitioners when developing and testing approaches and systems for computational law.
	
\end{abstract}


\section{Introduction}\label{sec:intro} 

There is a surge observed in research and applications of computer science and artificial intelligence in the legal domain. The related term \textit{computational law} is commonly defined as ``\textit{the branch of Legal Informatics concerned with the representation of rule and regulations in computable form}" \citep{genesereth2022introduction}. The focus of an important percentage of related work on computational law is on automatic processing, generation, or understanding of legal documents \citep{kucuk2024exploiting}.

Recent advancements in artificial intelligence (AI), such as generative AI models, pre-trained language models (PLMs) or large language models (LLMs), and chatbots developed using such models, have also affected the domain of computational law, and this dramatic impact is also acknowledged by legal professionals \citep{goth2023lawyers}. Undoubtedly, annotated or unannotated datasets and benchmarks in digital form are required for legal AI studies on legal texts, in order to facilitate model training, and to ensure sound comparisons of different approaches to the problems pertaining to computational law. Significant research have been conducted on the application of deep learning approaches to various natural language processing (NLP) problems in the legal domain, including text classification, text summarization, information extraction, named entity recognition, information retrieval, legal judgment prediction, reading comprehension, question answering, and recommendation systems \citep{chalkidis2019deep, cui2023survey, fragkogiannis2023context, ujwal2024reasoning}. Similarly, semantic resources such as ontologies are important for building related AI-based, or automated legal systems.

In this paper, we review the related literature on the datasets, benchmarks, and ontologies for computational law. Considering the fast pace of the published papers in the domain of computational law, this review paper can significantly contribute to both current and potential researchers and practitioners of the domain. While reviewing the relevant work, we observe the terms ``\textit{dataset}'' and ``\textit{benchmark}'' may be used interchangeably by the authors of the published papers, yet, we utilize the terms in the titles of the corresponding papers when grouping related studies under their relevant sections of our paper. 

The rest of the paper is organized as follows: In the first section below, we review the studies specifically on legal datasets, the next section is devoted those studies on legal benchmarks, the following section covers the studies on legal ontologies, and the final section concludes our paper.

\section{Datasets}\label{sec:datasets} 

A multi-label text classification dataset which includes 57K legal documents in English, that are extracted from EUR-Lex platform\footnote{\url{https://eur-lex.europa.eu/homepage.html}} for the laws of the European Union (EU), is presented in \citep{chalkidis2019large}. The dataset is named \textbf{EURLEX57K} and is annotated with about 4.7K labels, the evaluation results of different neural models are tested on this dataset, and the evaluation results are also included \citep{chalkidis2019large}.

In \citep{leitner2020dataset}, a legal dataset including court decisions in German annotated for named entity recognition is presented. Along with the typical named entity classes of person, location, and organization; domain-specific class labels such as lawyer, judge, court, court decision, regulation, and contract are also considered during annotation. A total of 750 court decision documents are included in the dataset and 53,632 entities are annotated \citep{leitner2020dataset}.

We come across three datasets of Brazilian origin (hence, in Portuguese) within the context of computational law. The first dataset is called \textbf{LeNER-Br} and is annotated for the named entity recognition task \citep{luz2018lener}. This dataset contains 66 legal documents produced in courts and four legislation documents, thereby, a total of 70 documents. In addition to the named entity tags for person, location, organization names and temporal expressions, two domain-specific tags, one for laws and one for legal decisions, are also employed during the annotation of the dataset \citep{luz2018lener}. 

A Portuguese text summarization dataset, \textbf{RulingBR}, is described in \citep{de2018rulingbr}. This legal summarization dataset contains 10,623 Brazilian court decisions. 

Another dataset (called \textbf{VICTOR}) of Brazilian legal documents corresponding to appeals, annotated for document classification is presented \citep{de2020victor}. The dataset contains legal documents of 44,855 suits, amounting to 628,820 documents. The translations of document classes include initial judgment of the lower court, appeal petition, judgment, court order, among others \citep{de2020victor}.

In \citep{zheng2021does}, a large dataset (called \textbf{CaseHOLD}) of multiple-choice questions to determine \textit{holdings} in legal case documents, where \textit{holding} is a legal term denoting the governing rule for a set of facts in a case, during the application of law. The dataset contains a total of 53,137 questions, and the origin of the dataset is Harward Law Library case law corpus.

A question answering dataset for the legal domain of private international law is presented in \citep{sovrano2021dataset}. This dataset comprises three regulations of the European Parliament and of the Council in English, and a total of 17 relevant questions determined by legal experts.

Contract review in the legal domain is the analysis of contracts to reveal the rights and obligations of the parties of the contract, and a contract review dataset called \textbf{CUAD} is presented in \citep{hendrycks2021cuad}. The dataset includes 510 contracts belonging to 25 contract types, annotated for 41 label categories determined, and there are 13101 labeled clauses in the dataset after the annotation process.

\textbf{Pile of law} is a large open-source dataset of legal documents proposed in \citep{henderson2022pile}. It is claimed that this dataset is compiled from 35 sources which encompass court filings and opinions, legal analyses, regulations, and contracts, among others. The authors emphasize that this dataset can help inform related parties about the data filtering procedures that are required for responsible training of large language models \citep{henderson2022pile}.

\textbf{EUR-Lex-Sum} is a multi-lingual dataset composed of manually created summaries of legal documents in the aforementioned EUR-Lex platform of EU \citep{aumiller2022eur}. There exist document-summary pairs in EUR-Lex-Sum that are aligned for 24 languages of the union, and there are up to 1,500 document per language. The evaluation results of Lexrank extractive summarization algorithm \citep{erkan2004lexrank} on EUR-Lex-Sum dataset are also presented \citep{aumiller2022eur}.

A dataset for legal abstractive summarization, called \textbf{Multi-LexSum}, is presented in \citep{shen2022multi}. The dataset covers summaries created at different (multiple) granularity levels, ranging from one sentence summaries to large summaries with over 600 words. This dataset contained about 40K legal documents and 9K summaries, where these summaries are created by legal experts \citep{shen2022multi}.

The Cambridge Law Corpus (\textbf{CLC}) is a legal dataset of more than 250K cases compiled from the England and Wales courts of UK jurisdiction \citep{ostling2023cambridge}. The authors also create an annotated subset of 638 cases (from CLC) for case outcome extraction task, and present the evaluation results of RoBERTa- and GPT-based models for this task. The former models are fine-tuned on CLC while the latter ones are evaluated without fine-tuning. It is suggested that CLC can be used for benchmarking and fine-tuning of large language models for the legal domain \citep{ostling2023cambridge}.

A dataset of legal documents of Supreme Court of the US (SCOTUS) compiled from multiple sources (called \textbf{Super-SCOTUS}) is presented in \citep{fang2023super}. The data sources include Supreme Court DataBase (SCDB), and other online resources such as Wikipedia, Justia, Oyez, and SCOTUS websites. The dataset is used for decision direction prediction task which is a binary classification task into either conservative or liberal \citep{fang2023super}.

\textbf{MAUD}, short for Merger Agreement Understanding Dataset, is another legal dataset annotated for reading comprehension \citep{wang2023maud}. The dataset contains 152 public merger agreements in English and 47,457 annotations based on extracted text from these agreements. 8,226 of these annotations are deal point text annotations while 39,231 of them are question-answer annotations (also called examples) \citep{wang2023maud}.

Related work on datasets of computation law demonstrate that there is a large body of research and there are several datasets in Chinese. To begin with, \textbf{CAIL2018} (Chinese AI and Law 2018) is a Chinese large-scale dataset for judgment prediction, and contains 2,676,075 criminal cases, which is described together with the competition of the same name in \citep{xiao2018cail2018}. Each case sample in the dataset has two parts: a fact description and a judgment result, where the judgment result, in turn, encompasses law articles, charges, and prison terms. The number of distinct criminal law articles in the dataset is 183 while the number of distinct charges and prison terms is 202 \citep{xiao2018cail2018}. 

In a subsequent study \citep{xiao2019cail2019}, \textbf{CAIL2019-SCM} (short for Chinese AI and Law 2019 Similar Case Matching) dataset is described, together with the corresponding competition, where the dataset covers 8,964 triplets of the fact descriptions of Chinese cases. The problem of similar case matching is defined as the determination of the case, among the second and third cases, that is the most similar to the first case in each triplet of the dataset.

\textbf{JEC-QA} is a dataset of multiple-choice questions compiled from questions of National Judicial Examination in China and practice questions from the Internet \citep{zhong2020jec}. It contains 26,365 questions where they are either knowledge-driven or case analysis questions.

A dataset of Chinese cases for legal case retrieval called \textbf{LeCaRD} (Legal Case Retrieval Dataset) is presented in \citep{ma2021lecard}. The dataset contains a total of 107 query cases, and 43,823 criminal case documents, additionally, novel relevance judgment criteria are also proposed within the course of the study \citep{ma2021lecard}. An extended version of this dataset called \textbf{LeCaRDv2} which includes 800 query cases, and 55,192 case documents, and 50 charges \citep{li2024lecardv2}.

\begin{table}[h]
  \caption{Distribution of Event Types and Mentions in the LEVEN Dataset \citep{yao2022leven}}
    \label{fig:leven}
    \includegraphics[width=1.0\textwidth]{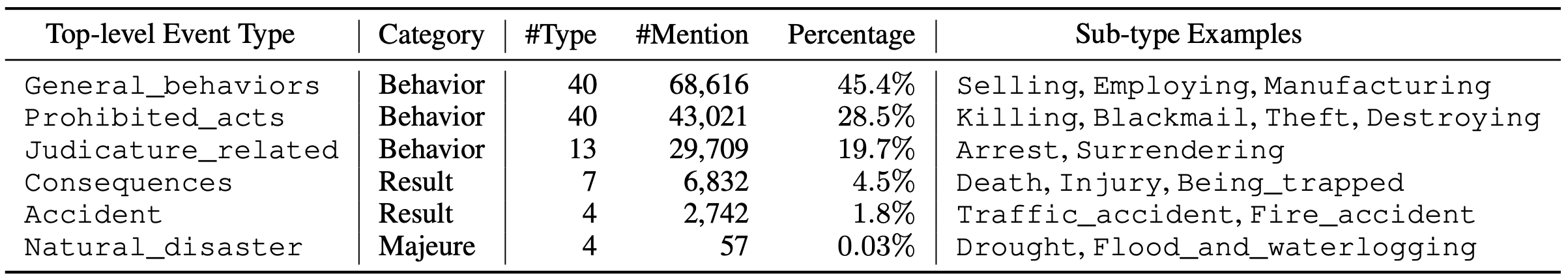}
\end{table}

After emphasizing the importance of determining the facts (in terms of events) before making judgments in the legal domain, an event detection dataset called \textbf{LEVEN} is proposed in \citep{yao2022leven} which includes 8,116 legal documents in Chinese which have been annotated with 108 event types, amounting to 150,977 event mentions in the dataset. The distribution of event types and mentions in LEVEN are presented in Table \ref{fig:leven}, as excerpted from \citep{yao2022leven}.

A large dataset (called Court Debate Dataset - \textbf{CDD}) of 30,481 Chinese private lending cases is described in \citep{ji2023cdd}. While the original dataset, which includes the dialogues between judges, plaintiffs, and defendants  is in Chinese, its English translation is also shared by the paper's authors. Experienced judges have contributed to the annotation procedure by defining the aspects of annotation, such as whether the litigation period has expired or not, whether the interest rate is agreed on, and whether the loan is a private loan. Additionally, other information such as loan amount, loan period, and interest payment are also annotated on the dataset. Annotations are performed by law school students and it is concluded that the dataset can be used for various problems in the legal domain including text classification, and summarization \citep{ji2023cdd}.

\textbf{EQUALS} (lEgal QUestion Answering via reading Chinese LawS) is a dataset comprising 6,914 triplets of law articles, questions, and answers, for legal question answering problem \citep{chen2023equals}. It is emphasized that spans of the law articles are annotated by senior law students to create the dataset. The dataset spans 10 Chinese laws including the civil code, civil procedure law, criminal law, and criminal procedure law \citep{chen2023equals}.

Similarly, \textbf{LeDQA} (short for Legal case Document-based Question Answering) is a dataset for legal question answering dataset comprising Chinese civil law case documents \citep{liu2024ledqa}. There are 100 cases, together with 4,800 case-question pairs, and 132,048 relevance annotations at the sentence level in this dataset. Within this context, the legal document-based question answering is defined as the task of generating the accurate answer, when a case document and a question about this case are given \citep{liu2024ledqa}.

Swiss Leading Decision Summarization (\textbf{SLDS}) dataset is introduced in \citep{rolshoven2024unlocking} for the legal text summarization task. It is suggested that SLDS is a multilingual legal dataset which includes 18,175 Swiss judicial decisions in German, French, and Italian, with their summaries produced by clerks and judges in German. The study emphasizes that the resource can be used both by researchers of legal text summarization and by legal practitioners \citep{rolshoven2024unlocking}.

In \citep{hou2024clerc}, \textbf{CLERC} (Case Law Evaluation and Retrieval Corpus) is described which is a dataset built for the tasks of legal case retrieval and retrieval-augmented generation. CLERC is based on the case law data compiled within the course of Caselaw Access Project of Harward Law School\footnote{\url{https://case.law/}} where this original data source contains more than 1.8 million federal case documents. The dataset is made up of CLERC/doc, CLERC/passage, and CLERC/queries for legal information retrieval, and CLERC/generation for for legal text generation \citep{hou2024clerc}.

In order to automate the process of determining the target actors, conditions, and exceptions in legal rules of laws, \textbf{LegalDiscourse} dataset is proposed \citep{spangher2024legaldiscourse}. The dataset includes more than 100,000 state laws, applicable throughout the US. These laws are annotated from a discourse analysis perspective, with 8 discourse elements including subject, object, consequence, and exception \citep{spangher2024legaldiscourse}.

\begin{figure}[h]
    \centering
    \includegraphics[width=0.79\textwidth]{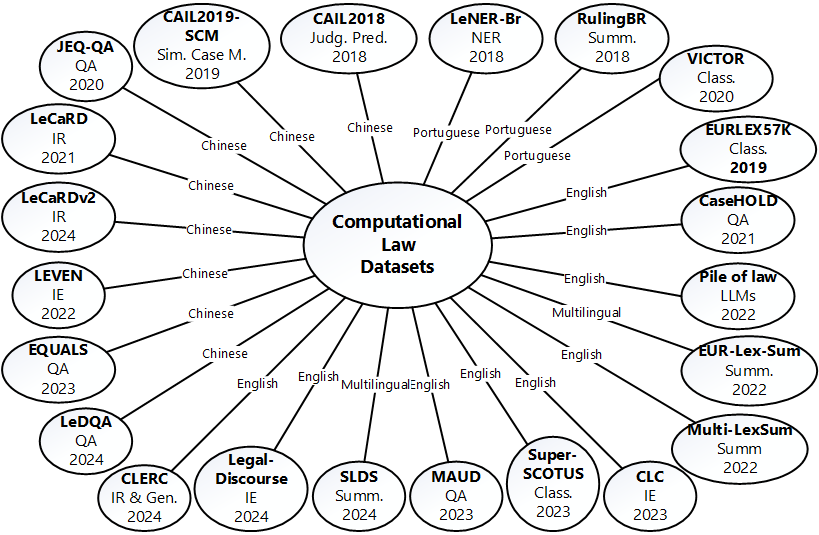}
    \caption{Named Datasets of Computational Law}
    \label{fig:sets}
\end{figure}

A schematic representation of only the named datasets of computational law covered so far is given in Figure \ref{fig:sets}. In the figure, in  each of the nodes denoting datasets; below the names of the datasets, an abbreviated form of the corresponding task or problem is given (such as `\textit{Summ.}' for summarization, `\textit{NER}' for named entity recognition, `\textit{QA}' for question answering, and `\textit{Sim. Case M.}' for similar case matching), and finally the dates of publications of the datasets are given below the abbreviations. The edges in the figure denote the languages of the corresponding datasets.

\section{Benchmarks}\label{sec:benchmarks} 

A legal benchmark called Swiss Judgment Prediction (\textbf{SJP}) including 85K legal cases (in German, French, and Italian) with dates between 2000 and 2020 from the Federal Supreme Court of Switzerand is described in \citep{niklaus2021swiss}. Along with this dataset proposed for the problem of legal judgment prediction, the evaluation results of several BERT-based models are also presented. It is observed that (language-specific) native BERT models perform better than multilingual BERT models on the SJP dataset \citep{niklaus2021swiss}.

\textbf{MILDSum} is a legal benchmark dataset for cross-lingual summarization of court decisions in English to summaries in English and Hindi \citep{datta2023mildsum}. The dataset contains about 3K legal decisions in English, which are summarized in English and Hindi by legal practitioners. Models from two different cross-lingual summarization approaches are employed and the corresponding results are reported. The first approach is monolingual summarization and then translation approach, while the second one is direct cross-lingual summarization \citep{datta2023mildsum}.

\textbf{LexGLUE} (Legal General Language Understanding Evaluation) is a benchmark dataset covering several significant legal datasets in English \citep{chalkidis2021lexglue}. LexGLUE covers seven legal datasets that encompass legal documents of the US Supreme Court, European Court of Human Rights, and from other corpora of US and EU law. Relevant NLP task for six of these seven datasets is text classification while the last dataset was created for multiple choice question answering. Various learning models including based on transformers and BERT, along with SVM classifiers are tested for these tasks on the datasets, and the best performing models are reported \citep{chalkidis2021lexglue}.

In \citep{hwang2022multi}, a legal AI benchmark in Korean, called \textbf{LBOX OPEN}, is presented. It includes a dataset of 147K cases, and covers several legal NLP tasks such as classification, legal judgment prediction, and summarization. The authors also propose a decoder-only legal language model for Korean (based on GPT-2 model), pre-trained on LBOX OPEN \citep{hwang2022multi}.

\textbf{FairLex} is a legal benchmark encompassing evaluation results pre-trained language models for fairness on four published legal datasets (with data from the European Court of Human Rights, US Supreme Court, Federal Supreme Court of Switzerland, and Supreme People’s Court of China) \citep{chalkidis2022fairlex}. The datasets include cases in five different languages, making FairLex a multilingual benchmark, and the considered NLP tasks include legal judgment prediction, and legal topic classification \citep{chalkidis2022fairlex}.

\begin{figure}[h]
    \centering
    \includegraphics[width=0.63\textwidth]{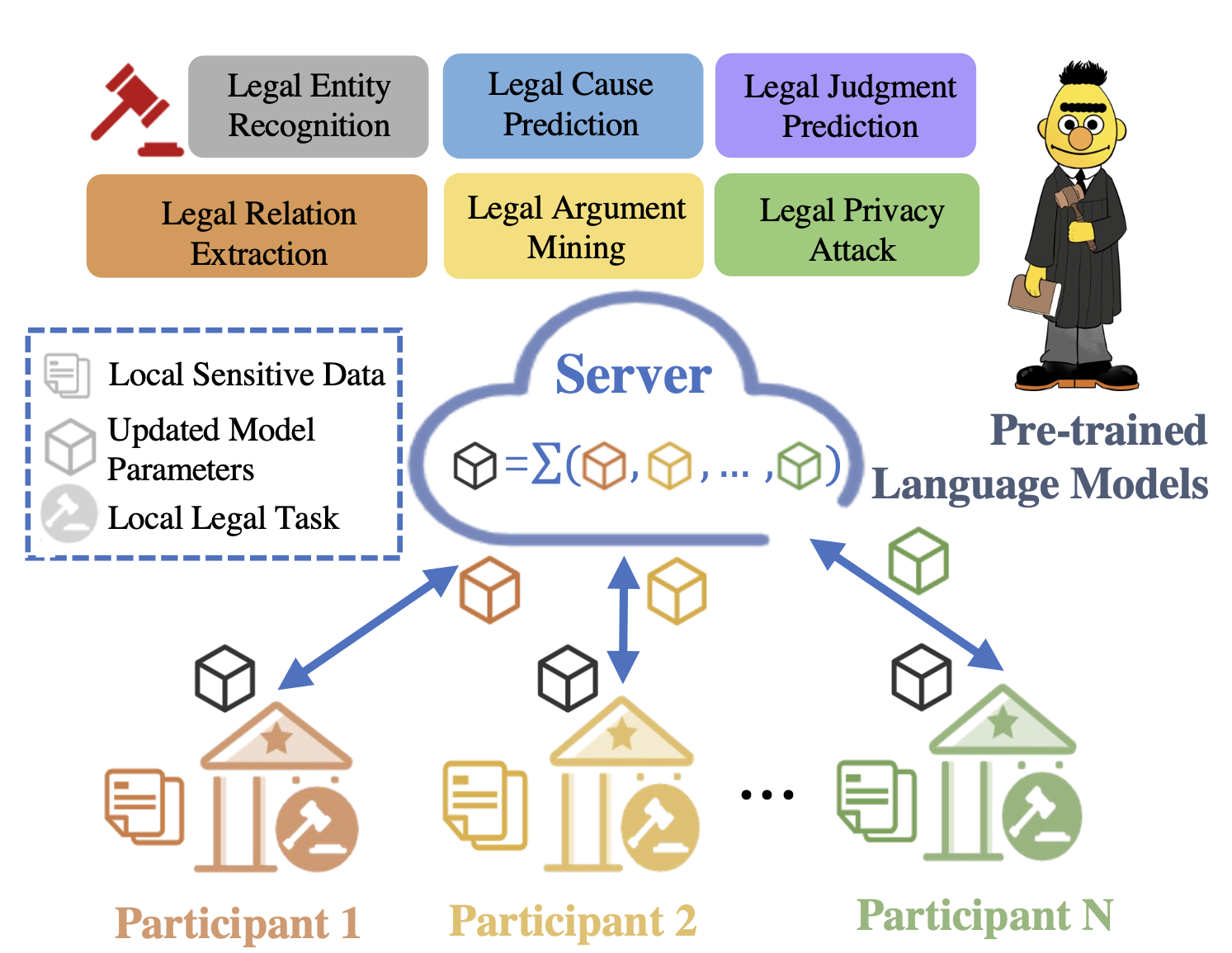}
    \caption{Schematic Overview of FEDLEGAL Benchmark (As excerpted from \citep{zhang2023fedlegal})}
    \label{fig:fedlegal}
\end{figure}

\textbf{FEDLEGAL} is proposed as a legal federated learning benchmark which aims to alleviate the negative effects of training learning models on sensitive legal documents \citep{zhang2023fedlegal}. The benchmark focuses on five NLP tasks: legal cause prediction, judgment prediction, entity recognition, relation extraction, and argument mining. The performance evaluation results of different federated learning models are also reported and compared \citep{zhang2023fedlegal}. Schematic overview of FEDLEGAL benchmark is presented in Figure \ref{fig:fedlegal}, as excerpted from the original study \citep{zhang2023fedlegal}.

\textbf{LegalBench} is a large legal benchmark that is collaboratively created to evaluate 162 different legal reasoning tasks in large language models (LLMs) \citep{guha2023legalbench}. The benchmark has a typology of the related reasoning tasks, and covers existing legal datasets in addition to new hand-crafted ones by legal professionals. The performance evaluations of 20 LLMs belonging to 11 families (such as versions of GPT, LLaMA, Claude, BLOOM, among others) and the corresponding results are reported \citep{guha2023legalbench}. In a subsequent study, LegalBench is extended as \textbf{LegalBench-RAG} for legal retrieval tasks \citep{pipitone2024legalbench} due to the effectiveness of Retrieval-Augmented Generation (RAG) models \citep{lewis2020retrieval}. This extended benchmark extracts relevant and minimal segments from the legal texts to improve retrieval precision. Different RAG pipelines are proposed and evaluated on the final benchmark \citep{pipitone2024legalbench}.

\textbf{LEXTREME} is proposed a multi-lingual benchmark which is composed of 11 legal datasets in a total of 24 languages, where the publication dates of these datasets range from 2010 to 2022 \citep{niklaus2023lextreme}. Five different encoder-based multilingual models are evaluated on LEXTREME, and their results are reported along with best performing monolingual models \citep{niklaus2023lextreme}.

\textbf{LAiW} is a Chinese benchmark for training large language models for different text processing tasks in the legal domain \citep{dai2023laiw}. The authors categorize different tasks into three groups as basic information retrieval (which includes tasks such as named entity recognition, recommendation, and summarization), legal foundation inference (covering tasks such as similar case matching and legal question answering), and finally, complex legal application (for tasks including judicial reasoning generation). Next, they combine the existing publicly available datasets for these subproblems together with propriety data, in order to build the LAiW benchmark \citep{dai2023laiw}.

Another legal benchmark for Chinese is \textbf{LexEval} which covers 23 tasks and 14,150 questions, using a new legal taxonomy called LexAbility Taxonomy \citep{li2024lexeval}. The LexEval benchmark includes both existing legal datasets, and manually-created ones. The evaluation results of a total of 38 LLMs are reported in the related study \citep{li2024lexeval}, where some of these LLMs are general and some of them are specific to the legal domain.

\textbf{ArabLegalEval} is a legal benchmark proposed for question answering (QA) tasks in Arabic legal documents \citep{hijazi2024arablegaleval}. A methodology for creating legal QA datasets for Arabic is also presented, along with the evaluation results of different LLMs on the underlying dataset.

\textbf{IL-TUR} is a legal benchmark for Indian legal texts which targets at eight NLP tasks, some of which are either for Hindi or English (hence, monolingual) while others are multilingual (for nine Indian languages) \citep{joshi2024tur}. The benchmark is based on documents from Indian jurisdiction and the related NLP tasks include text classification, summarization, machine translation, information extraction, and information retrieval. Various models based on BERT and GPT are tested on the datasets, and the results are presented and compared \citep{joshi2024tur}.

\begin{figure}[h]
    \centering
    \includegraphics[width=0.75\textwidth]{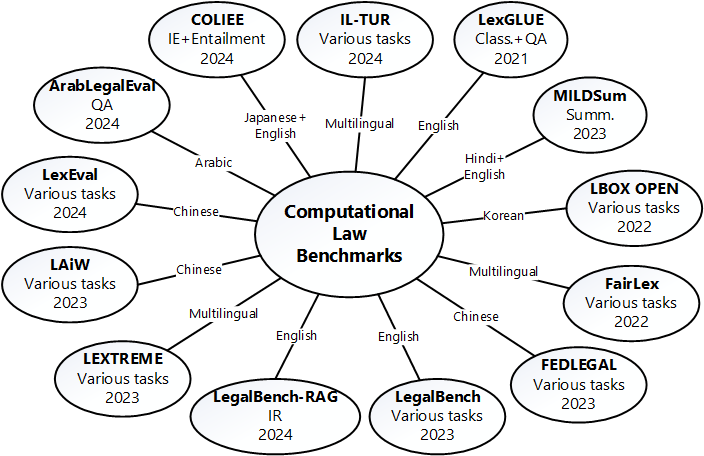}
    \caption{Named Benchmarks of Computational Law}
    \label{fig:benchmarks}
\end{figure}

The datasets provided within the course of the \textbf{COLIEE} (Competition on Legal Information Extraction/Entailment) are compiled from the judgments of Federal Court of Canada (in English), Japanese bar exam questions, and articles of Japanese civil codes \citep{goebel2024dataset}. As its name implies, the tasks considered are legal information extraction and legal textual entailment, both for case law and statute statute law.

A schematic summary of the covered benchmarks are demonstrated in Figure \ref{fig:benchmarks}. Similar to Figure \ref{fig:sets}; the names of the benchmarks, the tasks considered (as abbreviations), and the dates are given in the nodes, while the edges denote the languages of the corresponding benchmarks in Figure \ref{fig:benchmarks}.

\section{Ontologies}\label{sec:ontologies} 

Ontologies are used to conveniently represent domain information together with their structure, dependencies, and interrelationships \citep{uschold1996ontologies}. They serve various purposes including the organization of information, modelling, reasoning, semantic indexing and search, among others \citep{van2008ontologies}. For instance, well-known CYC ontology is a very large semantic resource of human commonsense knowledge, that has been built to facilitate information retrieval and integration \citep{lenat1995cyc}. Ontologies offer significant contributions to the legal domain as well, for instance, as underlying semantic resources for legal information retrieval, question answering, and recommendation systems to be built for the legal practitioners. Hence, this subsection is devoted a review of those studies on legal ontologies, i.e., on ontologies of computational law.

Related work reveals that legal ontologies have attracted the attention of the researchers and practitioners for decades. From the earliest to the latest ones, ontologies have been frequently used as semantic building blocks of larger legal information systems in the related studies \citep{mccarty1981representation, valente1999legal, veena2019ontology}.

One of the earliest studies dates back to 1981 and outlines a representation of concepts in the domain of corporate tax law with the ultimate intention to achieve an ontology of legal concepts in general \citep{mccarty1981representation}. Another study published in 1994 presents a functional ontology of law (\textbf{FOLaw}) which aims to include the specifications for main semantic concepts in law together with their organization \citep{valente1994functional}. The main knowledge categories for the concepts in this ontology are normative, responsibility, world, reactive, creative knowledge, and finally a sixth positional knowledge category which can be formed using the first five categories. Among these categories, normative knowledge category is considered as the core legal knowledge category and enables behaviors to be classified as either ``allowed" or ``disallowed" \citep{valente1994functional}. This core category and its subcategories are derived from the influential work of legal philosopher Kelsen who classifies norms as commanding, permitting, empowering, and derogating, with respect to their functions \citep{kelsen1991general}.

Based on a refined version of the functional legal ontology \citep{valente1994functional}, a legal data storage, reasoning, and retrieval architecture called ON-LINE (ONtology-based Legal INformation Environment) is presented in \citep{valente1999legal}. In a subsequent study, insights about this functional ontology and another legal ontology called \textbf{LRI-Core} are presented \citep{breuker2004legal}. It is claimed that the functional ontology is significant for its being a detailed epistemological framework and there is a need for a reusable foundational ontology for the legal domain. Hence, the authors propose LRI-Core ontology which has the following five top categories: physical, mental, and abstract concepts, roles, and terms of occurrences. LRI-Core ontology is further used as the upper ontology of \textbf{CRIME.NL} which is an ontology specifically built for Dutch criminal law \citep{breuker2004legal}. Aforementioned two ontologies (FOLaw \citep{valente1994functional} and LRI-Core \citep{breuker2004legal}) are further elaborated in \citep{van2008ontologies} together with some insights about building ontologies for the legal domain. The authors claim that many of the legal concepts actually correspond to common-sense concepts and that building legal ontologies on existing core or foundational ontologies will lead to more useful and extensible legal ontologies with better expressive power \citep{van2008ontologies}.

A domain ontology for Italian criminal law is presented in \citep{asaro2003domain} where crime is the central concept which is composed of structural elements such as offender, behaviour, event, circumstances, and punishment. The ontology encompasses further sub-concepts for the aforementioned and applicable concepts, for instance, punishment can be life imprisonment, imprisonment, arrest, or fine. The author of \citep{lame2004using} presents the construction of a legal ontology from French codes, based on the assumption that ``\emph{the legislator, while making the law, conceptualizes the legal field}". Terms from a total of 57 Codes (available on the Internet) are analyzed to extract terms using a syntactic analyzer and next, the extracted terms are post-processed using statistical metrics to obtain an ultimate list of about 118,000 terms. Finally, using other syntactic and statistical methods, relations among the terms are identified. It is pointed out that the constructed legal ontology can be used in legal information retrieval systems \citep{lame2004using}.

A foundational ontology called Core Legal Ontology (\textbf{CLO}) and a lexical database of legal concepts called \textbf{JurWordNet} are presented in \citep{gangemi2005constructive}. It is claimed that both CLO and JurWordNet are populated with legal notions including law, legal norm, regulation, legal right, legal power, duty, privilege, legal role, legal agent, legal fact, and crime. Similar to the original lexicon WordNet \citep{miller1995wordnet}, JurWordNet is an ontology-driven semantic resource for the legal domain which can be used in semantic indexing, information extraction, and information retrieval systems \citep{gangemi2005constructive}.

An automatic legal ontology construction procedure using legal texts is presented in \citep{lenci2007nlp}. The authors employ an automatic tool called T2K (Text-to-Knowledge) that has NLP, statistical learning and machine learning capabilities which is applied on Italian legislative texts to automatically extract domain terms, structure them as taxonomical chains, and cluster related terms, as a case study \citep{lenci2007nlp}. Legal Knowledge Interchange Format (LKIF) and \textbf{LKIF core ontology} for the legal domain are presented at the same venue by \citep{hoekstra2007lkif}. LKIF is proposed to facilitate conversions between knowledge bases created using distinct formalisms and also to serve as a convenient knowledge representation formalism. LKIF core ontology is created to represent the basic concepts in the legal domain including those related to norms, actions, agents, time, and place \citep{hoekstra2007lkif}.

In a related book published in 2011 devoted to engineering of legal ontologies, it is emphasized that legal ontologies lie at the intersection of the disciplines of law, artificial intelligence, and knowledge management, and the development of a legal ontology called Ontology of Professional Judicial Knowledge (\textbf{OPJK}) is presented as a case stuy \citep{casellas2011legal}. In line with previous research, it is pointed out that legal ontologies play an important role in the development of systems for legal information representation, information retrieval, reasoning and argumentation \citep{casellas2011legal}.

\begin{figure}[h]
    \centering
    \includegraphics[width=0.75\textwidth]{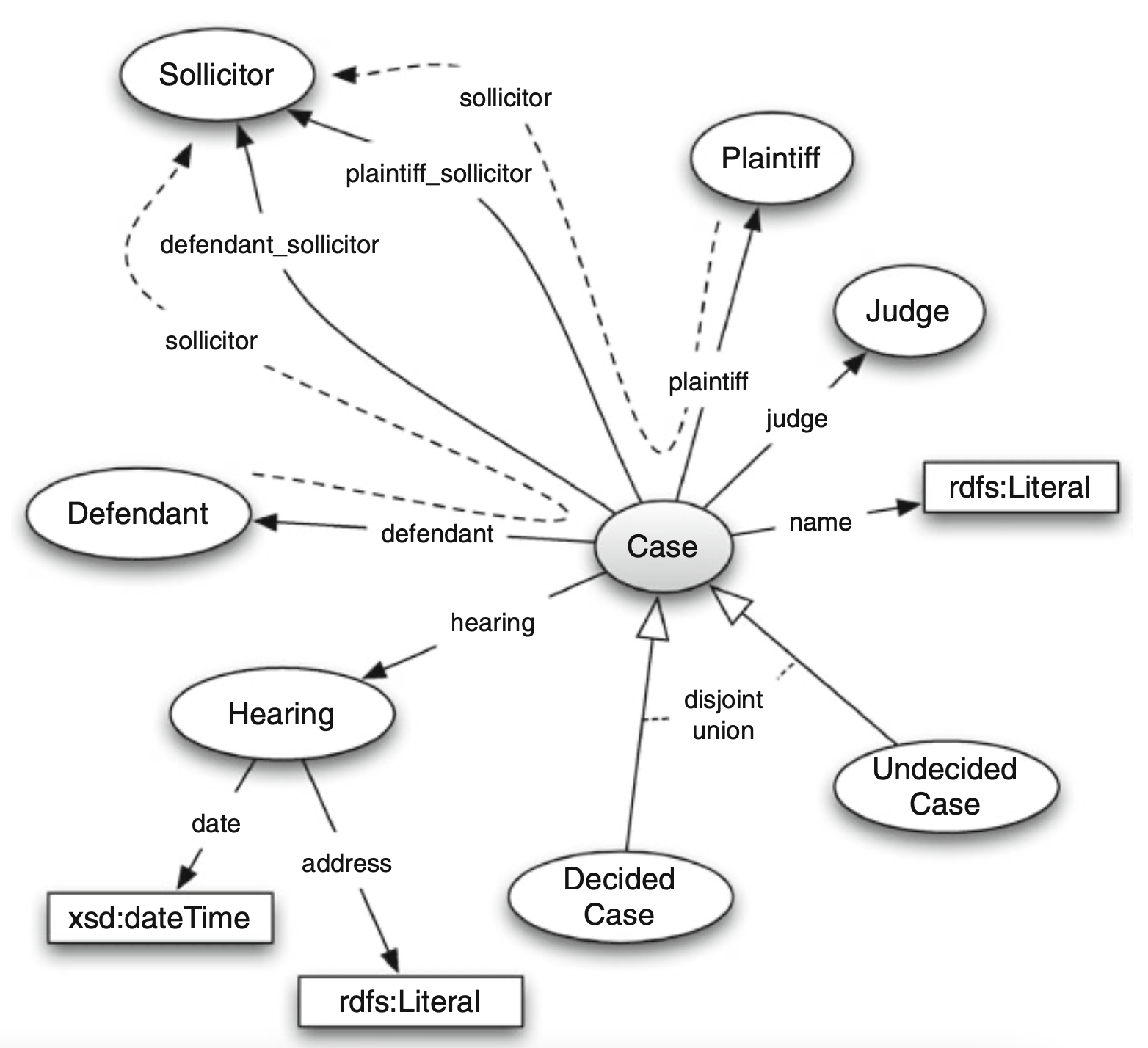}
    \caption{\textit{Case} Concept from the Legal Case Ontology \citep{wyner2012legal}}
    \label{fig:case}
\end{figure}

A legal case ontology was proposed in \citep{wyner2012legal} with an instantiation of a genuine and unusual case (of property law) about the ownership of a famous baseball, where the case is handled by California Superior Court and the court made a decision regarding this case in 2002. To illustrate, the \textit{Case} concept from this legal case ontology is demonstrated in Figure \ref{fig:case} together with its relationships and related concepts \citep{wyner2012legal}.

The \textbf{JudO} ontology is presented in \citep{ceci2016owl} particularly for the judicial decision representation and reasoning using the Web Ontology Language (OWL version 2). It is defined as composed of two modules: a core ontology that is aligned with the aforementioned LKIF core ontology \citep{hoekstra2007lkif} and a domain ontology which is particularly proposed for Italian Consumer Code \citep{ceci2016owl}.

A legal ontology specifically constructed for the data protection by modelling the related concepts and norms in General Data Protection Regulation (GDPR), which is applicable in all European Union (EU) countries, is presented in \citep{palmirani2018legal} where the ultimate data privacy-related ontology is named as \textbf{PrOnto}. It is claimed that PrOnto conveniently models the concepts in GDPR including data types, documents, roles, agents, purposes, rules, and deontic operators. The deontic operators (related to the duties and violations) include violation, compliance, prohibition, and obligation. It is suggested that the goal of building PrOnto is to utilize it for automatic legal reasoning and compliance checking \citep{palmirani2018legal}.

A systematic mapping study regarding legal ontologies is presented in \citep{de2019legal} where the related studies presented after being categorized with respect to the purpose of the ontologies presented, their generality levels, and the legal theories that they are based on. It is claimed that reusability of legal ontologies is still an open issue that should be improved with further studies. Similar to this study, an analysis of the state-of-the-art regarding legal ontologies is presented in \citep{leone2020taking} where the considered ontologies are classified with respect to their domains such as policies, privacy, licences, and tenders and procurements.

A question answering system based on the Motor Vehicle Act in India, that utilizes an ontology created for this legal subdomain, is presented in \citep{veena2019ontology}. In order to create the ontology, the authors use semantic role labeling to determine the actors, offenses, and penalties while processing online documents related to the aforementioned act. Questions posed to the system are similarly analyzed and the legal ontology is used as a semantic resource to answer the questions \citep{veena2019ontology}.

A core legal ontology specifically proposed for Vietnamese legal documents is described in \citep{nguyen2022vilo} where this ontology is given the name of \textbf{ViLO}. It is pointed out that the ontology covers concepts within Vietnamese political system and legal documents, and that it can be used by further AI applications in the legal domain. Another legal ontology called \textbf{Legal-Onto} is also proposed and used to build an information retrieval system particularly for Vietnamese Land Law \citep{nguyen2022legal}.

Regarding EU regulations, a lightweight multilingual ontology which aims to represent the legal terminology on EU Consumer Law, is proposed in \citep{ajani2016european}, where it is called European Legal Taxonomy Syllabus (\textbf{ELTS}). It is emphasized that ELTS encompasses two sections: a legal concept repository and a legal terminological database of related linguistic patterns. Another lightweight ontology for EU criminal procedures rights, mainly based on six EU directives on safeguarding people during criminal proceedings and investigations, is presented in \citep{audrito2023analogical}. The ontology includes related definitions, and is made up of concepts (representing principles) linked to each other with purpose links.

\begin{figure}[h]
    \centering
    \includegraphics[width=0.75\textwidth]{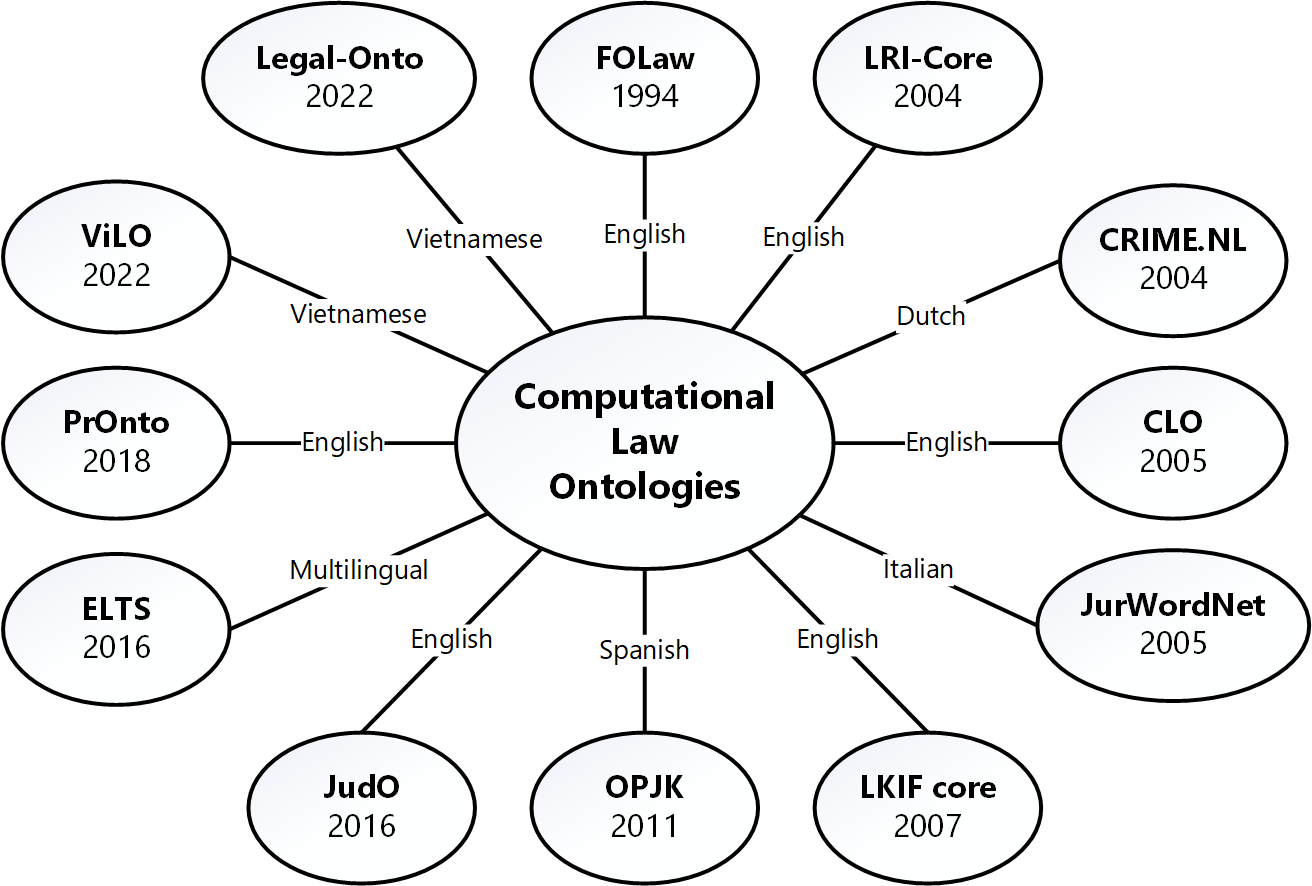}
    \caption{Named Ontologies of Computational Law}
    \label{fig:onto}
\end{figure}

We conclude this subsection on domain ontologies in computational law with a schematic representation of the named ontologies that we have reviewed, given in Figure \ref{fig:onto}. The dates below the ontology names correspond to the publication years of the papers that we cite in the current paper. Nevertheless, we should note that these dates may not be strictly accurate as the ontologies might have actually been proposed in earlier publications of their authors. On the edges (in the figure) connecting the ontologies to the center, we display the language of each ontology, as understood from its corresponding paper.

\section{Conclusion}\label{sec:conc} 

Shared datasets and benchmarks for computational law are invaluable for training and fine-tuning machine\slash deep learning models, and large language models for the legal domain. Apart from such resources, ontologies are important particularly for building practical computational legal systems. Therefore, the focus of the current paper is to review related studies on datasets, benchmarks, and ontologies for computational law. We believe that this up-to-date review is a significant contribution to related body of knowledge, as it compiles and presents significant resources for researchers and practitioners of computational law.

\bibliographystyle{unsrtnat}

\end{document}